%% file: main.tex
\documentclass[letterpaper]{article}
\input{sec/header}

\title{Decoupled Contrastive Learning for Federated Learning}
\author{
    Hyungbin Kim,
    Incheol Baek,
    Yon Dohn Chung
}
\affiliations{
    Korea University \\
    \{
    hyungbinkim,
    inch307,
    ydchung
    \}@korea.ac.kr
}

\begin{document}
\maketitle

\input{sec/00-abstract}
\input{sec/01-intro}
\input{sec/02-related}
\input{sec/03-preliminaries}
\input{sec/04-method}
\input{sec/05-experiment}
\input{sec/06-conclusion}

\clearpage
\bibliography{main}

\clearpage
\appendix
\input{sec/07-supp}
\end{document}

%% file: sec/header.tex
\usepackage{aaai2026}  
\usepackage{times}  
\usepackage{helvet}  
\usepackage{courier}  
\usepackage[hyphens]{url}  
\usepackage{graphicx} 
\usepackage{natbib}  
\usepackage{caption} 
\usepackage{algorithm, algorithmic}
\usepackage{booktabs}
\usepackage{amsmath}
\usepackage{graphicx}
\usepackage{subcaption}
\usepackage{amssymb}
\usepackage{breqn}
\usepackage{amsthm}
\usepackage{multirow}
\usepackage{makecell}
\usepackage{placeins}

\newtheoremstyle{plain}
  {8pt}{8pt}
  {\itshape}        
  {}                    
  {\bfseries}            
  {.}                    
  { }                    
  {}                     

\theoremstyle{plain}

\newtheorem{theorem}{Theorem}

\newtheorem{proposition}{Proposition}

\usepackage{newfloat}
\usepackage{listings}
\DeclareCaptionStyle{ruled}{labelfont=normalfont,labelsep=colon,strut=off} 
\floatstyle{ruled}
\newfloat{listing}{tb}{lst}{}
\floatname{listing}{Listing}
\nocopyright

%% file: sec/00-abstract.tex
\begin{abstract}
Federated learning is a distributed machine learning paradigm that allows multiple participants to train a shared model by exchanging model updates instead of their raw data. However, its performance is degraded compared to centralized approaches due to data heterogeneity across clients. While contrastive learning has emerged as a promising approach to mitigate this, our theoretical analysis reveals a fundamental conflict: its asymptotic assumptions of an infinite number of negative samples are violated in finite-sample regime of federated learning. To address this issue, we introduce \textit{Decoupled Contrastive Learning for Federated Learning (DCFL)}, a novel framework that decouples the existing contrastive loss into two objectives. Decoupling the loss into its alignment and uniformity components enables the independent calibration of the attraction and repulsion forces without relying on the asymptotic assumptions. This strategy provides a contrastive learning method suitable for federated learning environments where each client has a small amount of data. Our experimental results show that DCFL achieves stronger alignment between positive samples and greater uniformity between negative samples compared to existing contrastive learning methods. Furthermore, experimental results on standard benchmarks, including CIFAR-10, CIFAR-100, and Tiny-ImageNet, demonstrate that DCFL consistently outperforms state-of-the-art federated learning methods.
\end{abstract}

%% file: sec/01-intro.tex
\section{Introduction}
Federated Learning (FL) is a distributed learning framework that enables a shared model to be trained across distributed devices (termed \textit{clients}) without sharing of raw data of clients \cite{mcmahan2017communication}. By exchanging only model updates rather than raw data, FL mitigates privacy risks. However, since clients train on locally collected data, the data distribution across them is often not independent and identically distributed (non-IID). This gives rise to the primary challenges of data heterogeneity among clients and class imbalance within each local dataset. Under non-IID settings, these heterogeneity and imbalance cause a misalignment between the global and local objectives. This phenomenon—commonly referred to in the FL literature as \textit{client drift}—leads to degraded model performance and slower convergence \cite{li2019convergence}.

The approaches to these challenges have focused on two complementary strategies: (1) stabilizing training during the local update phase \cite{li2020federated, lee2024fedsol, xu2025federated, fu2025virtual}, and (2) improving efficiency in the aggregation phase \cite{reddi2020adaptive, kim2024communication, xia2025multisfl, liu2025semidfl, huang2025pa3fed}. Recently, contrastive learning–based regularization methods \cite{li2021model, seo2024relaxed, mu2023fedproc} have been introduced to constrain local models from deviating excessively. The objective of these methods is to maximize the cosine similarity between the representations of an anchor and a positive sample (from the same class), while minimizing the cosine similarity to negative samples (from different classes). In contrastive learning, the mechanism that aligns the representations of a positive pair is termed the \textit{attraction} force, while the mechanism separating them from negative samples is termed the \textit{repulsion} force. The intuition behind using contrastive learning is that by encouraging the model to learn generalizable and robust feature representations on local clients, the negative impact of data heterogeneity can be mitigated during global model aggregation.

However, these methods rely on the implicit assumption that standard contrastive objectives are directly applicable to FL setting, neglecting the detail that these objectives were not designed for the inherent constraints of FL, specifically the finite number of samples per client. We demonstrate that applying standard contrastive learning to the FL setting is theoretically unsuitable. Existing works show that a larger number of negative samples leads to better downstream task performances \cite{wang2020understanding, chen2020simple, awasthi2022more}. In \cite{wang2020understanding}, researchers show that standard contrastive learning operates under the asymptotic assumption of infinite negative samples, which allows its loss function to be decomposed into an \textit{alignment} term, which encourages the features of positive pairs to be similar, and a \textit{uniformity} term, which promotes a uniform distribution of features from negative samples. Furthermore, their analysis indicates that the standard contrastive loss approaches its theoretical optimum as the number of negative samples becomes exceedingly large. This poses a fundamental conflict with FL environments, where each client's dataset is small and finite. Therefore, it is essential to develop a contrastive learning approach tailored to the unique constraints of FL.

In this paper, we propose \textit{Decoupled Contrastive Learning for Federated Learning (DCFL)}, a novel framework for robust representation learning in FL. Unlike existing contrastive regularization methods that rely on a single contrastive loss, DCFL decouples the forces of attraction and repulsion into two distinct objectives, alignment and uniformity. This design enables independent control over each force via two hyperparameters, $\lambda_a$ and $\lambda_u$. Furthermore, our method offers both sample-wise and prototype-wise contrastive learning under a unified formulation. Our contributions are summarized as follows:

\begin{itemize}
    \item We first provide a theoretical analysis that demonstrates the incompatibility between the standard contrastive loss and the finite-sample constraints of FL.
    \item We propose \textit{Decoupled Contrastive Learning for Federated Learning (DCFL)}, a novel framework that separates the loss into its alignment and uniformity objectives. Our non-asymptotic design allows for the independent calibration of attractive and repulsive forces.
    \item DCFL is designed for versatility, unifying both sample-wise and prototype-wise contrastive learning approaches under a single, coherent formulation.
    \item Evaluation results on multiple standard federated learning datasets demonstrate that our method consistently outperforms state-of-the-art federated learning methods.
\end{itemize}

%% file: sec/02-related.tex
\section{Related Work}
In this section, we overview federated learning approaches from two complementary perspectives: (1) local training, which focuses on how clients optimize model parameters under heterogeneity constraints, and (2) global aggregation, which addresses strategies for effectively combining client updates into a robust global model. Additionally, we discuss recent approaches that integrate contrastive learning into federated learning frameworks, leveraging representation learning techniques to further mitigate data heterogeneity and improve model generalization.

\subsection{Federated Learning}
Federated Learning (FL) has emerged as an effective framework for collaborative model training across distributed clients without compromising data privacy \cite{mcmahan2017communication}. FL consists of a central server and multiple participants, who train a model on their local data and then send only the resulting model updates to the server for aggregation. This approach provides the advantages of preserving data privacy and reducing the computational cost on the server, as models are updated locally by the participants. However, data heterogeneity across distributed clients leads to significant challenges, such as slower convergence and performance degradation compared to traditional centralized learning. To overcome these limitations, existing studies in FL have mainly focused on two complementary perspectives: local training and global aggregation.

Researchers have focused on local training methods to enhance model convergence and mitigate client drift. FedProx \cite{li2020federated} introduces a proximal term in local objectives to stabilize training against heterogeneous client updates. Similarly, SCAFFOLD \cite{karimireddy2020scaffold} employs control variates to correct local updates, reducing client drift. FedDecorr \cite{shi2022towards} shows that data heterogeneity leads to dimensional collapse in representations and introduces a regularization term during local training to enforce decorrelation among feature dimensions. FedSOL \cite{lee2024fedsol} finds a robust parameter space that balances the conflicting objectives of global alignment and local generality. Our framework is a client-side approach that specifically focuses on reducing client drift among heterogeneous local clients and enhancing generalization performance.

In addition to local training methods, server-side aggregation strategies have been studied to address client heterogeneity through update normalization \cite{wang2020tackling}, layer-wise alignment \cite{wang2020federated, mendieta2022local}, and adaptive weighting \cite{reddi2020adaptive, kim2024communication, xia2025multisfl}. As our framework is based on the client-side approach, it is orthogonal to server-side techniques and can be easily combined with these methods to further improve performance.

\subsection{Contrastive Learning in FL}
Contrastive learning has been increasingly integrated into FL to tackle the challenges posed by heterogeneous data and client drift. These works can be classified into three complementary categories—sample-level \cite{dong2021federated, collins2021exploiting, seo2024relaxed}, model-level \cite{li2021model}, and prototypical contrastive methods \cite{mu2023fedproc, tan2022federated}—each aiming to align local representations and stabilize training while preserving privacy and imposing only modest additional communication overhead.

FedRCL \cite{seo2024relaxed}, a recent sample-level method, adaptively imposes the divergence penalty on pairs of examples in the same class and prevents their representations from being learned to be indistinguishable. MOON \cite{li2021model}, which is a pioneering model-level method, maximizes similarity between local and global model embeddings while repelling from the client’s previous state, correcting drift caused by skewed local updates. Prototypical contrastive strategies introduce shared class prototypes—either computed on the server or aggregated from clients—as anchors for local feature alignment. FedProc \cite{mu2023fedproc} enforces a contrastive loss between local embeddings and these global prototypes, narrowing the distribution gap among clients and improving generalization.

\textbf{Positioning Our Work.} While these studies differ in their approach, they share the implicit assumption that standard contrastive objectives are directly applicable to the FL setting. Motivated by this foundation, we demonstrate that applying standard contrastive learning to the FL setting is theoretically unsuitable and propose a novel contrastive loss function suitable for supervised learning scenarios under heterogeneous data in FL.

%% file: sec/03-preliminaries.tex
\section{Preliminaries}

\subsection{Federated Learning}

We consider a federated learning (FL) setting consisting of $N$ clients, each possessing a local dataset $\mathcal{D}_k = \{(\mathbf{x}_i, y_i)\}_{i=1}^{n_k},$ where $x_i \in \mathbb{R}^d$ is an input sample and $y_i \in \{1, \dots, C\}$ is the corresponding label. Let $n_k = |\mathcal{D}_k|$ denote the number of samples at client $k$, and let $n = \sum_{k=1}^N n_k$ be the total number of samples across all clients. The objective of FL is to collaboratively optimize a global model $f_{\theta}$ parameterized by $\theta$ without directly sharing local datasets, thereby maintaining data privacy.

Formally, the FL objective function is defined as:
\begin{dmath}
\min_{\theta}\ \mathcal{L}(\theta)
\quad\text{with}\quad
\mathcal{L}(\theta)
= \sum_{k=1}^N \frac{n_k}{n}\,\mathcal{L}(\theta^{k}),
\end{dmath}
where the local objective at client $k$ is
\begin{dmath}
\mathcal{L}(\theta^{k})
= \frac{1}{n_k}\sum_{(\mathbf{x},y)\in\mathcal{D}_k}\ell\bigl(f_{\theta}(\mathbf{x}),\,y\bigr),
\end{dmath}
and $\ell(\cdot,\cdot)$ denotes the cross-entropy loss. Training proceeds in communication rounds: at each round, clients perform local updates on $\theta$ and send model updates to a central server, which aggregates them (e.g., by Federated Averaging \cite{mcmahan2017communication}).

\subsection{Supervised Contrastive Learning}

Supervised contrastive learning extends the contrastive loss paradigm by leveraging label information to enhance the discriminative power of learned representations. Given a batch of samples, the supervised contrastive loss aims to bring representations of samples from the same class closer while pushing apart representations of samples from different classes.

Formally, the supervised contrastive loss \cite{khosla2020supervised} is defined as:
\begin{dmath}
\mathcal{L}_{\mathrm{SupCon}} = -\sum_{j \neq i, \ y_j=y_i} \log \frac{\exp(\mathrm{sim}(z_i,z_j)/\tau)}{\sum_{k \neq i}\exp(\mathrm{sim}(z_i,z_k)/\tau)},
\label{supcon}
\end{dmath}
where $z_i = h(f_\theta(\mathbf{x}_i))$ is the normalized embedding of sample $\mathbf{x}_i$, $h(\cdot)$ denotes the representation extractor, $\text{sim}(\cdot,\cdot)$ indicates cosine similarity, $\tau$ is the temperature hyperparameter.

Integrating supervised contrastive learning into FL scenarios aims to mitigate representation drift among clients due to heterogeneous data distributions, thus improving overall global model robustness.

%% file: sec/04-method.tex
\section{Proposed Method}

\subsection{Rationale for a Decoupled Contrastive Learning Approach in Federated Learning}
\label{sec:theory}
The theoretical foundations of our \textit{Decoupled Supervised Contrastive Learning for Federated Learning (DCFL)} framework are derived from the \textit{alignment} and \textit{uniformity} principles for representation learning first formalized by \cite{wang2020understanding}. As mentioned in the introduction, the standard contrastive loss is unsuitable for the finite-sample nature of FL. To address this, we first examine how its underlying asymptotic assumptions break down. We then introduce DCFL, which resolves this issue by explicitly decoupling the loss function. This provides a principled solution for the finite-sample regime under the heterogeneous data distributions common in FL.

\subsubsection{Breakdown of asymptotic assumptions in FL.}
\label{ssec:breakdown}

Prior work has demonstrated that a larger number of negative samples leads to better performance in contrastive learning \cite{wu2018unsupervised, he2020momentum, tian2020contrastive}. In the limit of infinite negative samples ($M \to \infty$), the loss decomposes into two objectives: one optimizing for the alignment of similar samples and the other for the uniformity of the feature distribution.

Let \(f\) be an encoder that maps an input to its feature representation. The temperature parameter is denoted by \(\tau\), and \(M\) is the number of negative samples. A positive pair \((\mathbf{x}, y)\) is drawn from the distribution \(p_{\text{pos}}\), while the anchor \(\mathbf{x}\) and negative samples \(\mathbf{x}^{-}\) are drawn from the global data distribution \(p_{\text{data}}\). The following theorem provides a formal justification, demonstrating that the optimization of the limiting loss is driven by the dual objectives of alignment and uniformity.

\begin{theorem}[Asymptotics of \(\mathcal{L}_{\mathrm{SupCon}}\) \cite{wang2020understanding}]
\label{theor:objectives}
For a fixed \(\tau > 0\), as the number of negative samples \(M \to \infty\), the contrastive loss converges to:
\upshape
\begin{dmath}
    \lim_{M \to \infty} \left(\mathcal{L}_{\mathrm{SupCon}}(f; \tau, M) - \log M \right) = 
-\frac{1}{\tau} \mathbb{E}_{(\mathbf{x},y) \sim p_{\text{pos}}} \left[ f(\mathbf{x})^\top f(y) \right]
+ \mathbb{E}_{\mathbf{x} \sim p_{data}} \left[ \log \mathbb{E}_{\mathbf{x}^{-} \sim p_{\text{data}}} \left[ e^{{f(\mathbf{x}^{-})^\top f(\mathbf{x})}/{\tau}} \right] \right].
\end{dmath}
\end{theorem}

\begin{proposition}[Finite-Sample Error Bound \cite{wang2020understanding}]
\label{prop:error_bound}
The deviation between the actual contrastive loss and its asymptotic limit is bounded as follows:
\begin{equation}
\begin{split}
& \left|  \lim_{M\to\infty} \left(\mathcal{L}_{\mathrm{SupCon}}(f; \tau, M) - \log M \right) \right. \\
& \qquad \left. - (\mathcal{L}_{\mathrm{SupCon}}(f; \tau, M) - \log M) \right| \\
\le & \;\frac{1}{M} e^{2/\tau} + \mathcal{O}(M^{-1/2}).
\end{split}
\end{equation}
\end{proposition}

The error term, which arises from the difference between the true expected similarity over the global data distribution and the empirical average calculated from the finite \(M\) negative samples, decays slowly at a rate of $\mathcal{O}(M^{-1/2})$. This slow convergence poses a fundamental challenge in FL settings. Each client in an FL environment operates on a limited local dataset, meaning the number of available negative samples \(M\) is small. With a small $M$, the $\mathcal{O}(M^{-1/2})$ error term becomes large, causing the actual loss function experienced by the client to deviate from the theoretically ideal decomposition.

Therefore, the foundational asymptotic assumptions of standard contrastive loss are violated in the practical, finite-sample regime of FL. The loss function optimized by each client deviates substantially from the ideal global objectives of alignment and uniformity. This fundamental incompatibility motivates the development of a non-asymptotic approach.

\subsection{Decoupled Supervised Contrastive Learning}
\subsubsection{Detailed decomposition.}
We rewrite Eq.~(\ref{supcon}) by separating the numerator and denominator inside the logarithm using log properties:
\begin{dmath}
\mathcal{L}_{\mathrm{SupCon}} = - \sum_{j \neq i, \ y_j = y_i} \left[ \log \exp\left(\frac{\mathrm{sim}(z_i,z_j)}{\tau}\right) - \log\sum_{k \neq i}\exp\left(\frac{\mathrm{sim}(z_i,z_k)}{\tau}\right) \right] \
= - \sum_{j \neq i, \ y_j = y_i}\frac{\mathrm{sim}(z_i,z_j)}{\tau} + \sum_{j \neq i, \ y_j = y_i}\log\sum_{k \neq i}\exp\left(\frac{\mathrm{sim}(z_i,z_k)}{\tau}\right).
\label{eq:log_decompose}
\end{dmath}

Since the second term does not depend on the index $j$ explicitly, it can be simplified as:
\begin{dmath}
\sum_{j \neq i, \ y_j = y_i} \log\sum_{k \neq i}\exp\left(\frac{\mathrm{sim}(z_i,z_k)}{\tau}\right) = |P_i| \log\sum_{k \neq i}\exp\left(\frac{\mathrm{sim}(z_i,z_k)}{\tau}\right),
\end{dmath}
where $P_i = { z_j : y_j = y_i, j \neq i }$ denotes the set of positive samples for anchor $z_i$. Thus, we have:
\begin{dmath}
\mathcal{L}_{\mathrm{SupCon}} = - \sum_{p\in P_i}\frac{\mathrm{sim}(z_i,p)}{\tau} + |P_i|\log\sum_{k \neq i}\exp\left(\frac{\mathrm{sim}(z_i,z_k)}{\tau}\right).
\label{eq:supcon}
\end{dmath}

\subsubsection{Decoupling alignment and uniformity.}
To fully decouple alignment and uniformity terms, we exclude positive samples from the uniformity term. This modification decouples the loss, allowing the uniformity term to independently enforce feature uniformity by pushing the anchor away from all available negative samples. Introducing the scaling hyperparameters $\lambda_a,\lambda_u\in (0,1)$ with $\lambda_a+\lambda_u=1$, we obtain the decoupled contrastive loss:
\begin{dmath}
\mathcal{L}_{\mathrm{DCFL}} = -\lambda_a \sum_{p\in P_i}\frac{\mathrm{sim}(z_i,p)}{\tau} + \lambda_u |P_i|\log\sum_{n\in N_i}\exp\left(\frac{\mathrm{sim}(z_i,n)}{\tau}\right),
\label{eq:dcl}
\end{dmath}
where $N_i = { z_k : y_k \neq y_i }$ denotes the negative samples. This ensures a clear separation and independent optimization of each term.

\begin{algorithm}[t]
\caption{DCFL}
\label{alg:dcl}
\begin{algorithmic}[1]
\REQUIRE communication rounds \(\{T\}\), batch embeddings \(\{z_i\}\), prototypes \(\{c_\gamma\}\) (for prototype-wise), temperature \(\{\tau\}\), hyperparameters \(\{\lambda_a,\lambda_u\}\), learning rate \(\{\eta\}\).

\vspace{0.5em}

\STATE \textbf{Server executes:}
\STATE Initialize \(\theta_{0}\)
\FOR{\(t=1,2,...,T\)} 
    \STATE Randomly sample \(K_t \subseteq K\)
    \STATE Send \(\theta_{t-1}\) to \(K_t\)
    \FOR{each client in \(K_t\)}
        \STATE $\theta_{t}^{k} \leftarrow \theta_{t-1}$
        \STATE \textbf{Client updates \(\theta_{t}^{k}\)}
        \STATE Client sends \(\theta_{t}^{k}\) to the server
    \ENDFOR
    \vspace{0.3em}
    \STATE $\theta_{t} = \frac{1}{K_t}\sum_{K_t \subseteq K} \theta_{t}^{k}$
    \vspace{0.3em}
\ENDFOR

\vspace{0.5em}

\STATE \textbf{Clients update \(\theta_{t}^{k}\):}
\FOR{each anchor embedding \(z_i\)}
  \STATE Construct positive samples \(P_i\), negative samples \(N_i\)
  \STATE Compute $\mathcal{L}_{\mathrm{DCFL}}$ using Eq.~(\ref{eq:dcl})
  \STATE $\mathcal{L} = \mathcal{L}_{\mathrm{CE}} + \mu\mathcal{L}_{\mathrm{DCFL}}$
  \STATE Update parameters: \(\theta_{t}^{k}\gets\theta_{t-1}-\eta\nabla_{\theta_{t-1}} \mathcal{L}\)
\ENDFOR

\end{algorithmic}
\end{algorithm}

\subsubsection{Decoupling alignment and uniformity in finite samples.}
To overcome the limitations of asymptotic approach, our proposed DCFL replaces the standard loss with two separate, non-asymptotic objectives designed specifically for finite-sample environments.

\begin{dmath}
\mathcal{L}_{\mathrm{DCFL}} = \underbrace{-\lambda_a \sum_{p\in P_i}\frac{\mathrm{sim}(z_i,p)}{\tau}}_{\text{Direct Alignment}} \underbrace{+ \lambda_u |P_i|\log\sum_{n\in N_i}\exp\left(\frac{\mathrm{sim}(z_i,n)}{\tau}\right)}_{\text{Local Uniformity}},
\label{eq:dcl}
\end{dmath}
This formulation is theoretically justified as it directly optimizes the core principles within the practical constraints of FL. The alignment term isolates and preserves the often-weak attractive force from the scarce positive samples available locally. The uniformity term enforces feature separation based on the available---finite and biased---local negative samples. The hyperparameters $\lambda_a$ and $\lambda_u$ provide a necessary mechanism to control the trade-off between these two terms.

By enforcing this separation, DCFL provides a robust theoretical foundation for contrastive learning in the practical FL environment. For more theoretical discussion, please refer to the supplementary material.

\begin{table*}[t]
  \small
  \centering
  \begin{tabular*}{\textwidth}{@{\extracolsep{\fill}} l *{12}{c} @{}}
    \toprule
    & \multicolumn{6}{c}{\textbf{CIFAR-10}} & \multicolumn{6}{c}{\textbf{CIFAR-100}} \\
    \cmidrule(lr){2-7} \cmidrule(lr){8-13}
    \textbf{Method} & \multicolumn{2}{c}{IID} & \multicolumn{2}{c}{$\alpha=0.5$} & \multicolumn{2}{c}{$\alpha=0.3$}
                    & \multicolumn{2}{c}{IID} & \multicolumn{2}{c}{$\alpha=0.5$} & \multicolumn{2}{c}{$\alpha=0.3$} \\
    & MAX & EMA & MAX & EMA & MAX & EMA
      & MAX & EMA & MAX & EMA & MAX & EMA \\
    \midrule
    FedAvg    & 90.47         & 90.12         & 90.09         & 89.17         & 89.47         & 87.43
              & 63.34         & 62.99         & 63.86         & 63.33         & 63.06         & 62.63       \\
    FedProx   & 90.50         & 90.20         & 90.21         & \underline{89.27} & 89.25     & 87.06        
              & 63.93         & 63.71         & 63.37         & 63.24         & 63.43         & 63.41       \\
    MOON      & 91.18         & 90.66         & 88.88         & 87.97         & 87.71         & 86.23        
              & 63.52         & 63.13         & 63.12         & 62.75         & 63.59         & 63.13       \\
    FedDecorr & 90.36         & 90.17         & 89.48         & 88.72         & 89.20         & 87.56        
              & 63.27         & 62.97         & 62.64         & 62.40         & 62.57         & 62.16       \\
    FedRCL    & 89.32         & 88.86         & 88.44         & 86.71         & 87.02         & 83.97        
              & 63.22         & 62.94         & 62.35         & 62.23         & 61.53         & 60.83       \\
    FedSOL    & 91.22         & 91.04         & 89.69         & 88.84         & 88.19         & 86.69        
              & 65.47         & 65.15         & 64.23         & 64.06         & 63.50         & 62.96 \\
    FedSCL    & 88.92         & 88.70         & 85.47         & 83.40         & 83.89         & 81.67        
              & 61.44         & 61.23         & 59.41         & 58.91         & 58.41         & 57.56       \\
    FedProc   & 91.10         & 90.99         & 90.39         & 87.95         & \textbf{90.15} & \textbf{88.84} 
              & 64.76         & 64.26         & \underline{64.92} & 64.39     & 64.38         & 64.02       \\
    \midrule
    \textbf{DCFL-SW (ours)} 
              & \underline{91.53} & \underline{91.30} & \textbf{90.68} & 89.16 & 89.66  & 87.79 
              & \underline{66.05} & \underline{65.82} & 64.88         & \underline{64.57} & \underline{65.25}  & \underline{64.10} \\
    \textbf{DCFL-PW (ours)} 
              & \textbf{91.72} & \textbf{91.39} & \underline{90.41} & \textbf{89.51} & \underline{89.74} & \underline{88.42} 
              & \textbf{66.38} & \textbf{65.85} & \textbf{65.57} & \textbf{65.54} & \textbf{65.54} & \textbf{64.49} \\
    \bottomrule
  \end{tabular*}
  \caption{Performance comparison from 5\% participation rate over 100 clients on CIFAR-10 and CIFAR-100 under different data heterogeneity ($\alpha$) settings. Also, we report exponential moving average accuracy with the parameter set to 0.9.}
  \label{tab:fl-performance-002}
\end{table*}

\subsubsection{Sample-wise and prototype-wise variants.}
Eq.~(\ref{eq:dcl}) generalizes naturally to two important cases:

\begin{itemize}
    \item \textbf{Sample-wise DCFL}: Positive samples \(P_i=\{z_i^+\}\), negative samples \(N_i=\{z_j^-:j\neq i\}\).
    \item \textbf{Prototype-wise DCFL}: Positive prototype \(P_i=\{c_{y_i}\}\), negative prototypes \(N_i=\{c_\gamma:\gamma\neq y_i\}\).
\end{itemize}

This versatility allows the proposed method to adapt seamlessly to diverse scenarios. The training procedure is summarized in Algorithm~\ref{alg:dcl}.

%% file: sec/05-experiment.tex
\section{Experiments}
\subsection{Experimental Setup}
\subsubsection{Datasets.}
We conducted experiments on three datasets to explore the impact of our method, including CIFAR-10 \cite{krizhevsky2009learning}, CIFAR-100 \cite{krizhevsky2009learning}, and Tiny-ImageNet \cite{le2015tiny}. To generate the training and test datasets, we employed the Dirichlet distribution $Dir(\alpha)$ \cite{hsu2019measuring}, with a symmetric parameter $\alpha \in \{0.3, 0.5, \infty\}$.

\subsubsection{Baselines.}
We evaluate the performance of our proposed DCFL under two settings—sample-wise (DCFL-SW) and prototype-wise (DCFL-PW). We then compare DCFL with several state-of-the-art federated learning methods, including FedAvg \cite{mcmahan2017communication}, FedProx \cite{li2020federated}, MOON \cite{li2021model}, FedDecorr \cite{shi2022towards}, FedProc \cite{mu2023fedproc}, FedRCL \cite{seo2024relaxed}, FedSOL \cite{lee2024fedsol}, and FedSCL. Here, FedSCL refers to a technique that applies sample-wise contrastive learning in a federated setting and has been employed as a baseline in prior work \cite{seo2024relaxed}.

\subsubsection{Implementation details.}
We considered a federated learning setting with 100 clients. We adopted sample fractions of 0.02, 0.05, and 0.1; unless otherwise specified, a sample fraction of 0.05 was used. Each client trained a local ResNet-18 backbone \cite{he2016deep} for 5 epochs using stochastic gradient descent with an initial learning rate of 0.01, an exponential decay factor of 0.998, and a regularization factor of 0.0005. For methods that require a proximal or contrastive hyperparameter $\mu$, we tuned $\mu$ from \{0.001, 0.01, 0.1, 1, 5, 10\} and report the best value in the supplementary material. Unless otherwise noted, throughout all experiments we set $\lambda_{a}$ to 0.9 and $\lambda_{u}$ to 0.1, with a batch size of 64, and a temperature $\tau$ of 0.5. We used accuracy as an evaluation metric, reporting both the maximum accuracy achieved (MAX) and its exponential moving average (EMA) over rounds. Our experiments were implemented in Python 3.11.7 with PyTorch 2.6.0 under CUDA 12.1, and executed on 4 NVIDIA Tesla V100 GPUs.

\subsection{Experimental Results}

\subsubsection{Accuracy comparison.}
Table \ref{tab:fl-performance-002} shows the accuracy of all methods at a 5\% participation rate under varying degrees of data heterogeneity. A comparison of different FL methods reveals that our proposed DCFL achieves the highest performance across almost all settings. Specifically, FedProc, which is based on prototypical contrastive learning, already achieves high performance across almost all settings; in comparison, DCFL further yields significant gains by independently balancing the attraction and repulsion forces during prototype contrastive learning. These findings demonstrate the effectiveness of DCFL under different levels of data heterogeneity.

\subsubsection{Scalability on varying number of clients.}
\begin{table*}[t]
  \small
  \centering
  \begin{tabular*}{\textwidth}{@{\extracolsep{\fill}} l *{12}{c} @{}}
    \toprule
    & \multicolumn{6}{c}{\textbf{CIFAR-10}} & \multicolumn{6}{c}{\textbf{CIFAR-100}} \\
    \cmidrule(lr){2-7} \cmidrule(lr){8-13}
    \textbf{Method} & \multicolumn{2}{c}{IID} & \multicolumn{2}{c}{$\alpha=0.5$} & \multicolumn{2}{c}{$\alpha=0.3$}
                    & \multicolumn{2}{c}{IID} & \multicolumn{2}{c}{$\alpha=0.5$} & \multicolumn{2}{c}{$\alpha=0.3$} \\
    & MAX & EMA & MAX & EMA & MAX & EMA
      & MAX & EMA & MAX & EMA & MAX & EMA \\
    \midrule
    FedAvg    & 89.81 & 89.30 & 89.54 & 89.04 & 89.01 & 87.61
              & 61.67 & 61.57 & 61.80 & 61.57 & 61.99 & 61.58 \\
    FedProx   & 89.81 & 89.56 & 89.33 & 88.55 & 89.11 & 87.36
              & 61.69 & 61.37 & 61.73 & 61.68 & 61.77 & 61.10 \\
    MOON      & 91.23 & 90.81 & 89.55 & 88.87 & 87.94 & 86.36
              & 62.12 & 61.85 & 61.50 & 61.14 & 62.68 & 62.15 \\
    FedDecorr & 89.70 & 89.46 & 88.99 & 88.27 & 88.56 & 87.02
              & 60.98 & 60.51 & 60.88 & 60.56 & 60.70 & 60.68 \\
    FedRCL    & 89.46 & 88.96 & 88.10 & 87.33 & 86.73 & 84.11
              & 61.35 & 60.93 & 60.79 & 60.55 & 59.99 & 59.68 \\
    FedSOL    & \underline{91.38} & \underline{91.20} & 90.13 & \textbf{89.58} & 88.90 & \textbf{88.26}
              & \underline{64.28} & \underline{63.92} & \underline{63.33} & 63.21 & 63.05 & 62.88 \\
    FedSCL    & 89.33 & 88.95 & 86.41 & 85.30 & 84.96 & 82.93
              & 60.24 & 60.04 & 58.57 & 58.06 & 56.17 & 55.92 \\
    FedProc   & 91.07 & 90.84 & 90.11 & 89.11 & 89.44 & 87.42
              & 63.50 & 63.10 & 62.72 & 62.50 & 63.05 & 62.88 \\
    \midrule
    \textbf{DCFL-SW (ours)}
              & 90.98 & 90.79 & \textbf{90.47} & 89.27 & \textbf{89.75} & \underline{88.19}
              & 63.86 & 63.65 & 64.30 & \underline{63.92} & \textbf{64.14} & \underline{63.52} \\
    \textbf{DCFL-PW (ours)}
              & \textbf{91.70} & \textbf{91.50} & \underline{90.26} & \underline{89.35} & \underline{89.55} & 87.55
              & \textbf{65.90} & \textbf{65.31} & \textbf{64.53} & \textbf{63.93} & \underline{63.96} & \textbf{63.57} \\
    \bottomrule
  \end{tabular*}
  \caption{Performance comparison from 10\% participation rate over 100 clients on CIFAR-10 and CIFAR-100 under different data heterogeneity ($\alpha$) settings. Also, we report exponential moving average accuracy with the parameter set to 0.9.}
  \label{tab:fl-performance-004}
\end{table*}

To evaluate the scalability of our proposed method, we assessed its performance under environments with different numbers of participating clients per round. The results for a 10\% client participation rate are presented in Table \ref{tab:fl-performance-004}. We also conducted experiments with a 2\% participation rate. For brevity, the detailed results are provided in the supplementary material.

As presented in Table \ref{tab:fl-performance-004}, DCFL demonstrates strong scalability by consistently outperforming competing methods across different settings, confirming that its effectiveness is not limited to a specific client scale. This robustness is particularly evident in the most challenging scenarios, such as on the CIFAR-100 dataset with high data heterogeneity, where our method achieves a clear performance gain over strong baselines. This consistent performance advantage demonstrates that our decoupled loss function provides a robust optimization framework, making DCFL a scalable and practical solution for real-world federated networks.

\subsubsection{Robustness on complex dataset.}
\begin{table*}[t]
  \small
  \centering
  \begin{tabular*}{\textwidth}{@{\extracolsep{\fill}} l *{6}{c} @{}}
    \toprule
    & \multicolumn{6}{c}{\textbf{Tiny-ImageNet}} \\
    \cmidrule(lr){2-7}
    
    \textbf{Method} & \multicolumn{2}{c}{IID} & \multicolumn{2}{c}{$\alpha=0.5$} & \multicolumn{2}{c}{$\alpha=0.3$} \\
    & MAX & EMA & MAX & EMA & MAX & EMA \\
    \midrule
    FedAvg         & 50.36           & 49.97           & 47.29           & 46.76           & 46.43           & 45.45           \\
    FedProx        & 49.70           & 49.16           & 47.46           & 47.11           & 46.78           & 45.68           \\
    MOON           & 49.59           & 49.54           & 47.20           & 46.65           & 46.71           & 45.12           \\
    FedDecorr      & 49.34           & 48.58           & 47.73           & 46.94           & 45.97           & 45.17           \\
    FedRCL         & 49.60           & 49.01           & 46.63           & 45.75           & 46.09           & 45.27           \\
    FedSOL         & 49.88           & 49.25           & \underline{47.89}           & 47.19           & 47.17           & 45.96           \\
    FedSCL         & 48.98           & 47.78           & 45.47           & 44.83           & 44.10           & 43.32           \\
    FedProc        & \underline{50.82} & \underline{50.20} & 47.68 & \underline{47.25} & \underline{47.24} & \underline{46.27} \\
    \midrule
    \textbf{DCFL-SW (ours)} & 50.40           & 49.58           & 47.61           & 47.18           & 46.50           & 45.46  \\
    \textbf{DCFL-PW (ours)} & \textbf{51.45}  & \textbf{50.74}  & \textbf{48.80}  & \textbf{48.32}  & \textbf{47.46}  & \textbf{46.92}  \\
    \bottomrule
  \end{tabular*}
  \caption{Performance comparison from 5\% participation rate over 100 clients on Tiny-ImageNet under different data heterogeneity ($\alpha$) settings. Also, we report exponential moving average accuracy with the parameter set to 0.9.}
  \label{tab:fl-performance-005}
\end{table*}

Table \ref{tab:fl-performance-005} presents the experimental results on Tiny‑ImageNet. Compared to CIFAR‑10 and CIFAR‑100, Tiny‑ImageNet includes more classes and a larger total number of samples, allowing us to assess robustness in more challenging scenarios. Across varying degrees of data heterogeneity, DCFL consistently achieves promising performance improvements in every setting. While FedProc already achieves high accuracy in most settings, DCFL further outperforms it by decoupling the alignment and uniformity terms. These results demonstrate the robustness of DCFL in challenging federated learning scenarios.

\subsubsection{Influence of $\lambda_a$ and $\lambda_u$.}
\begin{table*}[t]
  \small
  \centering
  \begin{tabular*}{\textwidth}{@{\extracolsep{\fill}} l *{12}{c} @{}}
    \toprule
    & \multicolumn{6}{c}{\textbf{CIFAR-10}} & \multicolumn{6}{c}{\textbf{CIFAR-100}} \\
    \cmidrule(lr){2-7} \cmidrule(lr){8-13}
    
    \textbf{Method} & \multicolumn{2}{c}{IID} & \multicolumn{2}{c}{$\alpha=0.5$} & \multicolumn{2}{c}{$\alpha=0.3$}
                    & \multicolumn{2}{c}{IID} & \multicolumn{2}{c}{$\alpha=0.5$} & \multicolumn{2}{c}{$\alpha=0.3$} \\
    & MAX & EMA & MAX & EMA & MAX & EMA
      & MAX & EMA & MAX & EMA & MAX & EMA \\
    \midrule
    \makecell[l]{DCFL-SW\\($\lambda_a = 0.3,\ \lambda_u = 0.7$)}
              & 87.32         & 87.18         & 73.07         & 70.67         & 20.18         & 10.26
              & 61.57         & 61.02         & 56.13         & 55.04         & 52.95         & 52.09       \\
    \makecell[l]{DCFL-SW\\ ($\lambda_a = 0.9,\ \lambda_u = 0.1$)}
              & \textbf{91.30}         & \textbf{90.80}         & \textbf{90.38}         & \textbf{88.03}         & \textbf{86.84}         & \textbf{81.76}
              & \textbf{68.17}         & \textbf{67.56}         & \textbf{66.78}         & \textbf{65.35}         & \textbf{65.35}         & \textbf{64.09} \\
    \midrule
    \makecell[l]{DCFL-PW\\($\lambda_a = 0.3,\ \lambda_u = 0.7$)}
              & \textbf{91.72}        & \textbf{91.47}         & 87.87         & 86.69         & 85.45         & 81.12
              & \textbf{69.16}        & 67.61         & 66.19         & 64.15         & 63.13         & 61.75 \\
    \makecell[l]{DCFL-PW\\ ($\lambda_a = 0.9,\ \lambda_u = 0.1$)}
              & 91.50        & 91.16          & \textbf{89.83}         & \textbf{88.40}         & \textbf{86.99}         & \textbf{83.98} 
              & 68.64        & \textbf{68.11} & \textbf{67.32}         & \textbf{65.70}         & \textbf{65.26}         & \textbf{64.08} \\
    \bottomrule
  \end{tabular*}
  \caption{Ablation study with different $\lambda_a$ and $\lambda_u$.}
  \label{tab:fl-performance-006}
\end{table*}

Table \ref{tab:fl-performance-006} shows the sensitivity of the hyperparameters $\lambda_a$ and $\lambda_u$ on CIFAR‑10 and CIFAR‑100. To analyze the influence of $\lambda_a$ and $\lambda_u$, we conducted experiments for two scenarios under the constraint $\lambda_a,\lambda_u\in (0,1)$ with $\lambda_a+\lambda_u=1$: one with a large $\lambda_a$ and another with a large $\lambda_u$. In most settings, assigning a larger weight to the alignment term ($\lambda_a>\lambda_u$) yields superior performance. In particular, under more heterogeneous data distributions, the performance gap widens significantly, and when $\lambda_u$ is large, the model fails to converge (e.g., $\lambda_a=0.3, \lambda_u=0.7, \alpha=0.3$ on CIFAR‑10). These observations indicate both that high values of $\lambda_u$ induce unstable early‑phase learning and that increasing $\lambda_a$ can effectively stabilize training and mitigate these convergence failures.

\subsection{Discussion on Representation Quality via Cosine Similarity Analysis}
\begin{figure*}[t!]
  \centering
  \begin{subfigure}[b]{0.45\linewidth}
    \centering
    \includegraphics[width=\linewidth]{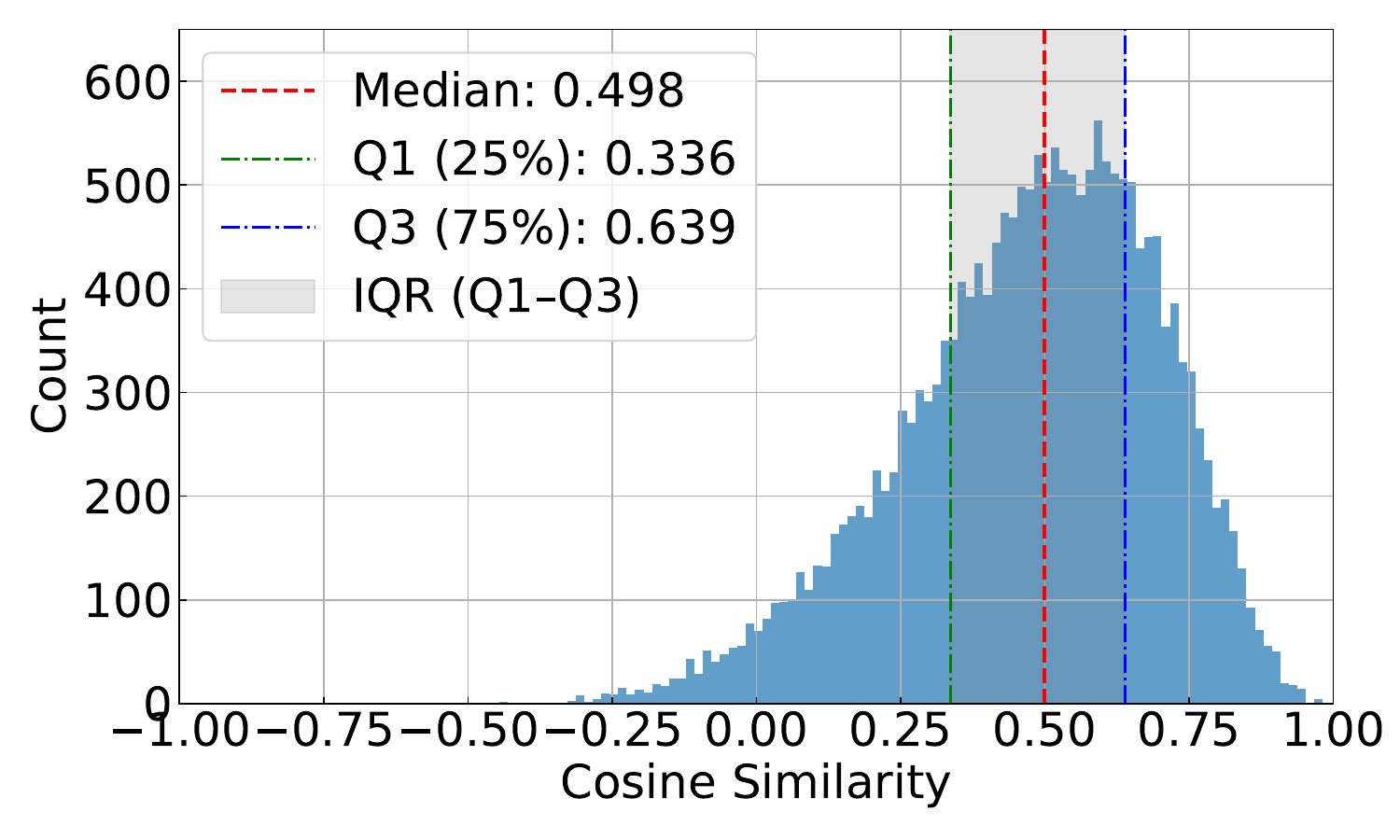}
    \caption{Intra‑class: FedProc}
    \label{fig:sub1}
  \end{subfigure}%
  \begin{subfigure}[b]{0.45\linewidth}
    \centering
    \includegraphics[width=\linewidth]{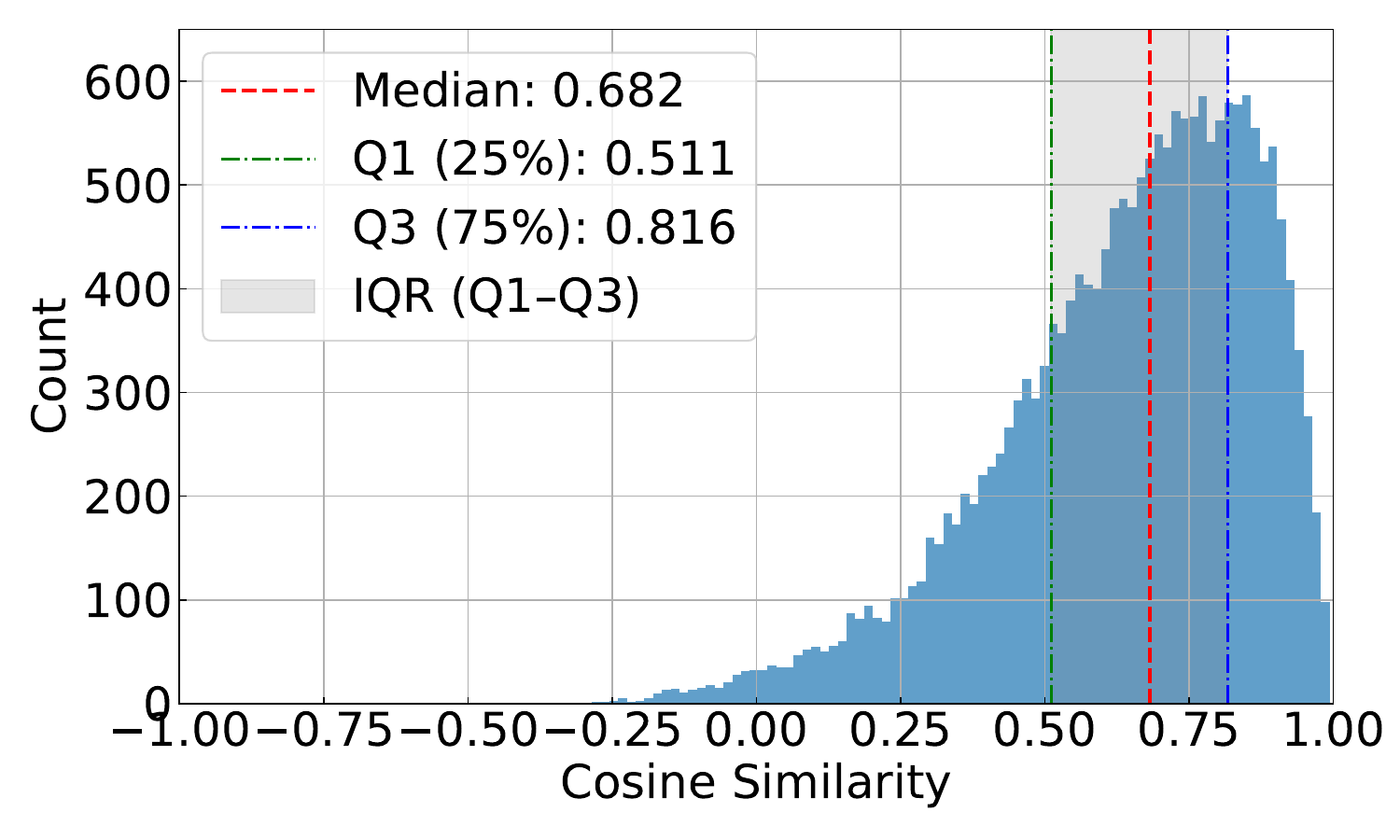}
    \caption{Intra‑class: DCFL-PW}
    \label{fig:sub2}
  \end{subfigure}
  
  \begin{subfigure}[b]{0.45\linewidth}
    \centering
    \includegraphics[width=\linewidth]{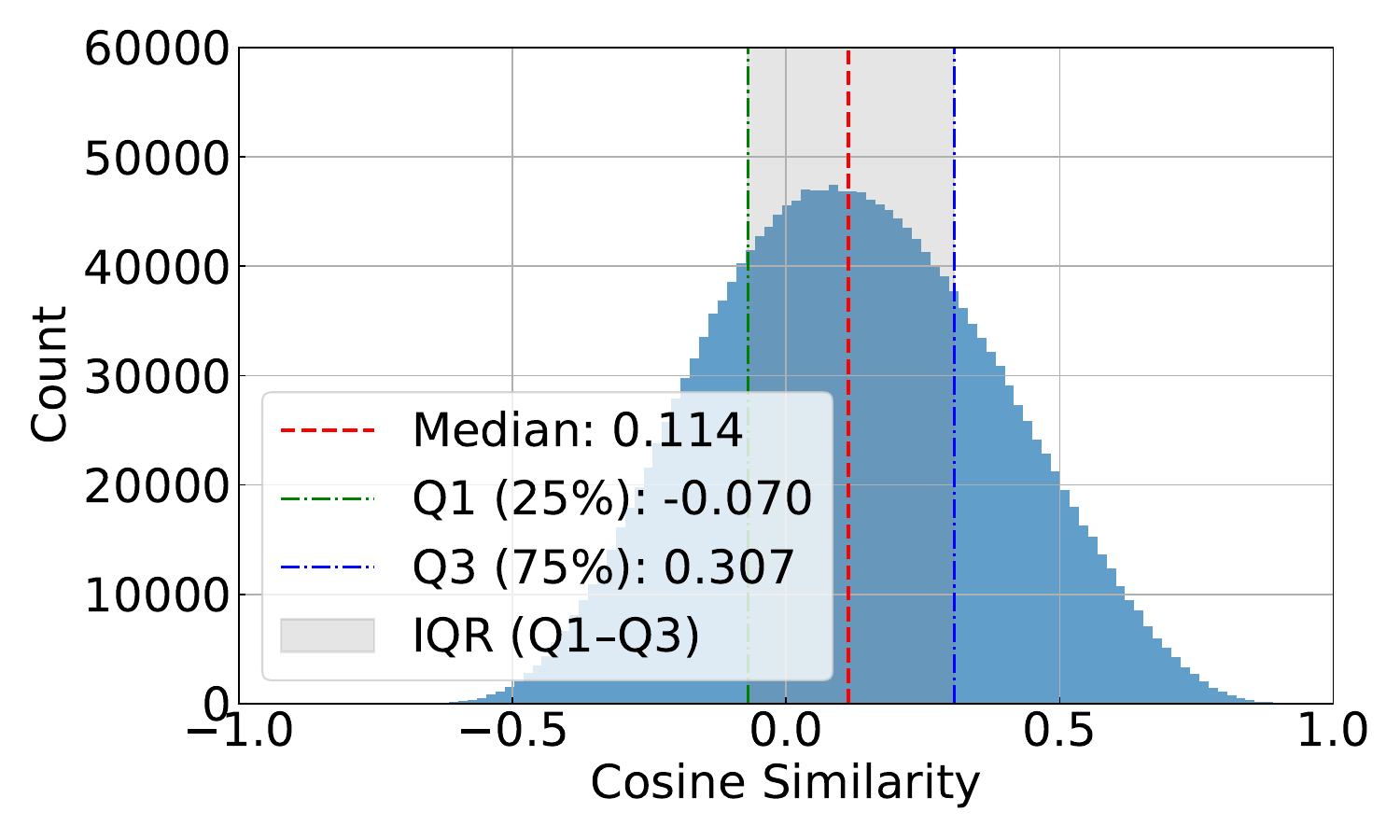}
    \caption{Inter‑class: FedProc}
    \label{fig:sub3}
  \end{subfigure}%
  \begin{subfigure}[b]{0.45\linewidth}
    \centering
    \includegraphics[width=\linewidth]{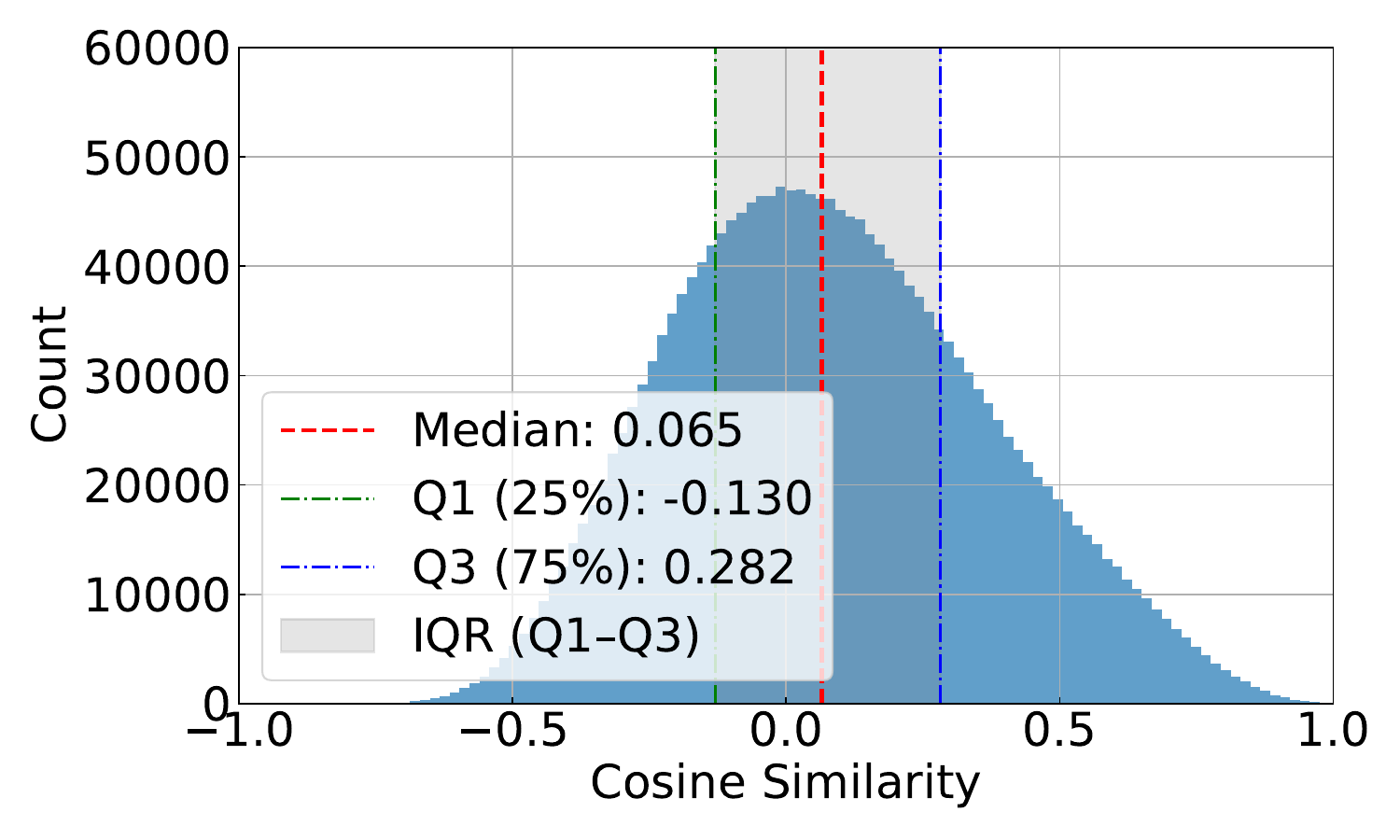}
    \caption{Inter‑class: DCFL-PW}
    \label{fig:sub4}
  \end{subfigure}

  \caption{Distribution of pairwise cosine similarities for samples, comparing FedProc and DCFL-PW.}
  \label{fig:cossim}
\end{figure*}

To demonstrate that DCFL successfully encourages both alignment and uniformity in an FL environment, we extracted the feature representations of test samples from two representative models, FedProc and our proposed DCFL-PW, and computed the pairwise cosine similarity for each model. We chose FedProc for comparison due to its robust performance among the baselines and its status as a prototypical contrastive learning method in FL. In this analysis, we consider two types of sample pairs: intra-class pairs (samples sharing the same label), for which cosine similarity values close to 1 indicate strong alignment in feature space; and inter-class pairs (samples from different labels), for which cosine similarity values near 0 indicate that representations are well separated and uniformly distributed.

Fig. \ref{fig:cossim} shows the intra‑ and inter‑class cosine similarity distributions for the DCFL model, revealing a clear separation between them. Specifically, the distribution of intra-class cosine similarities is tightly concentrated near 1, while the inter-class similarities concentrate around 0, demonstrating that DCFL effectively pulls together positives without collapsing or misplacing different-class samples. In contrast to DCFL, intra-class similarities of FedProc are more dispersed away from 1, and its inter-class similarities are less tightly clustered around 0, indicating weaker alignment and reduced uniformity.

%% file: sec/06-conclusion.tex
\section{Conclusion}
In this work, we identified and addressed a foundational conflict between theoretical assumptions of contrastive learning and the practical constraints of federated learning. Our analysis revealed how the reliance on infinite negative samples leads to unstable training in such a decentralized setting. To address this, we proposed \textit{Decoupled Contrastive Learning for Federated Learning (DCFL)}, a novel framework that separates the alignment and uniformity objectives, achieving effective representation learning. Our approach introduces hyperparameters to calibrate the alignment and uniformity terms, thereby enabling control over the attraction and repulsion forces. Experiment results confirmed that DCFL consistently outperforms state-of-the-art methods, particularly in challenging, highly heterogeneous environments. Furthermore, our visualizations demonstrate that DCFL outperforms existing federated contrastive learning methods in achieving both alignment and uniformity.

%% file: sec/07-supp.tex
\section*{Supplementary Material}

\section{Extended Theoretical Discussion}
We first review the principles of alignment and uniformity, then provide a deeper analysis of why the standard supervised contrastive loss is unsuitable for the Federated Learning (FL) environment.

\subsection{Revisiting Alignment and Uniformity}

As introduced by \cite{wang2020understanding}, the success of contrastive learning can be understood through two properties.
\begin{itemize}
    \item \textbf{Alignment:} Feature representations of similar samples (positive pairs) should be mapped to nearby points in the embedding space. This can be quantified by minimizing the expected distance between them.
    \begin{dmath}
        \mathcal{L}_{\mathrm{align}}(f; \alpha) \triangleq \mathbb{E}_{(\mathbf{x},y)\sim p_{\text{pos}}}[||f(\mathbf{x})-f(y)||_{2}^{\alpha}], \quad \alpha > 0.
    \end{dmath}

    \item \textbf{Uniformity:} The feature vectors of all samples should be distributed as uniformly as possible. This encourages the preservation of maximal information from the data. This can be optimized by minimizing the following loss, which is based on the average pairwise Gaussian potential:
    \begin{dmath}
        \mathcal{L}_{\mathrm{uniform}}(f; t) \triangleq \log \mathbb{E}_{\mathbf{x},y \stackrel{iid}{\sim} p_{\text{data}}}[e^{-t||f(\mathbf{x})-f(y)||_{2}^{2}}], \quad t > 0.
    \end{dmath}
\end{itemize}
The key insight from~\cite{wang2020understanding} is that the standard contrastive loss implicitly optimizes for both properties in asymptotic assumptions of an infinite number of negative samples.

\subsection{Analysis of the Coupled Loss in the FL Context}

The standard supervised contrastive loss ($\mathcal{L}_{\text{SupCon}}$) can be written as~\cite{khosla2020supervised}:
\begin{dmath}
\mathcal{L}_{\mathrm{SupCon}} = -\sum_{j \neq i, \ y_j=y_i} \log \frac{\exp(\mathrm{sim}(z_i,z_j)/\tau)}{\sum_{k \neq i}\exp(\mathrm{sim}(z_i,z_k)/\tau)},
\label{supcon}
\end{dmath}
where $z_i = h(f_\theta(\mathbf{x}_i))$ is the normalized embedding of sample $\mathbf{x}_i$, $h(\cdot)$ denotes the representation extractor, $\text{sim}(\cdot,\cdot)$ indicates cosine similarity, $\tau$ is the temperature hyperparameter.

The denominator term couples the attractive and repulsive forces. For any positive sample $z_j$, it contributes to the numerator (attraction) while also being part of the denominator's sum (repulsion).

In the IID setting, this formulation might lead to a stable training. However, in the FL setting where each client has a finite—and often small—amount of data, this structure is problematic.
\begin{itemize}
    \item The alignment signal, derived from the numerator, is inherently weak due to the scarcity of positive samples ($|P(i)|$ is small).
    \item The uniformity signal, derived from the denominator, is biased due to the non-IID nature of the negative samples.
\end{itemize}
The coupled loss forces a direct competition between this weak alignment signal and a biased uniformity signal. This makes it difficult for the model to learn meaningful representations. Our DCFL formulation (Eq. 10 in the main paper) solves this issue by separating the two forces, allowing each to be calibrated independently.

\section{Hyperparameter $\mu$ We Used}
\begin{table}[!htbp]
  \small
  \centering
  \begin{tabular}{lccc}
    \toprule
    Method   & CIFAR-10 & CIFAR-100 & Tiny-ImageNet \\
    \midrule
    MOON       & 5        & 1         & 1             \\
    FedProx    & 0.01     & 0.001     & 0.001         \\
    FedDecorr  & 0.01     & 0.01      & 0.01          \\
    FedProc    & 10       & 10        & 10            \\
    FedRCL     & 1        & 1         & 1             \\
    DCFL (ours) & 10      & 10        & 1            \\
    \bottomrule
  \end{tabular}
  \caption{Optimal $\mu$ values for different methods and datasets.}
  \label{tab:optimal-mu}
\end{table}

\section{Scalability on Varying Number of Clients}
\begin{table*}[htb!]
  \small
  \centering
  \begin{tabular*}{\textwidth}{@{\extracolsep{\fill}} l *{12}{c} @{}}
    \toprule
    & \multicolumn{6}{c}{\textbf{CIFAR-10}} & \multicolumn{6}{c}{\textbf{CIFAR-100}} \\
    \cmidrule(lr){2-7} \cmidrule(lr){8-13}
    \textbf{Method} &
    \multicolumn{2}{c}{IID} & \multicolumn{2}{c}{$\alpha=0.5$} & \multicolumn{2}{c}{$\alpha=0.3$} &
    \multicolumn{2}{c}{IID} & \multicolumn{2}{c}{$\alpha=0.5$} & \multicolumn{2}{c}{$\alpha=0.3$} \\
    & MAX & EMA & MAX & EMA & MAX & EMA
      & MAX & EMA & MAX & EMA & MAX & EMA \\
    \midrule
    FedAvg    & 91.06 & \underline{90.99} & 89.69 & \underline{89.06} & 86.16 & 78.46
              & 67.31 & 66.21 & 66.75 & 65.23 & 65.65 & 64.46 \\
    FedProx   & 91.13 & 90.64 & 89.96 & 89.02 & 86.31 & 75.16
              & 67.10 & 66.58 & 65.97 & 63.93 & 65.68 & 64.70 \\
    MOON      & 90.79 & 90.59 & 87.71 & 85.08 & 84.50 & 82.05
              & 67.20 & 66.35 & 65.95 & 64.50 & \textbf{65.91} & 64.57 \\
    FedDecorr & 91.22 & 90.80 & \underline{90.33} & \textbf{89.67} & 86.67 & 76.19
              & 66.15 & 65.56 & 66.01 & 64.15 & 65.64 & 64.53 \\
    FedRCL    & 90.32 & 90.00 & 88.06 & 86.80 & 84.74 & 76.74
              & 66.58 & 66.13 & 65.05 & 63.24 & 63.72 & 62.59 \\
    FedSOL    & 90.70 & 90.61 & 88.46 & 88.01 & 85.68 & \underline{83.37}
              & 66.72 & 66.25 & 65.39 & 64.66 & 65.32 & \textbf{65.19} \\
    FedSCL    & 88.61 & 88.58 & 85.41 & 84.39 & 81.14 & 75.62
              & 64.60 & 64.35 & 61.58 & 59.39 & 59.76 & 59.20 \\
    FedProc   & \underline{91.38} & 90.90 & 90.26 & 88.63 & 86.47 & 79.40
              & \underline{68.36} & \underline{67.69} & \underline{66.89} & 65.07 & \underline{65.83} & \underline{64.86} \\
    \midrule
    \textbf{DCFL-SW (ours)}
              & 91.30 & 90.80 & \textbf{90.38} & 88.03 & \underline{86.84} & 81.76
              & 68.17 & 67.56 & 66.78 & \underline{65.35} & 65.35 & 64.09 \\
    \textbf{DCFL-PW (ours)}
              & \textbf{91.50} & \textbf{91.16} & 89.83 & 88.40 & \textbf{86.99} & \textbf{83.98}
              & \textbf{68.64} & \textbf{68.11} & \textbf{67.32} & \textbf{65.70} & 65.26 & 64.08 \\
    \bottomrule
  \end{tabular*}
  \caption{Performance comparison from 2\% participation rate over 100 clients on CIFAR-10 and CIFAR-100 under different data heterogeneity ($\alpha$) settings. Also, we report exponential moving average accuracy with the parameter set to 0.9.}
  \label{tab:fl-performance-003}
\end{table*}

Table 2 in the main paper and Table \ref{tab:fl-performance-003} show that DCFL can scale to more challenging scenarios with varying numbers of participants. In practical FL settings, both data heterogeneity and participation rate heterogeneity play critical roles. We observe that DCFL consistently outperforms all other methods across almost all scenarios, regardless of variations in participation rate. Table \ref{tab:fl-performance-003} presents the robust performance gains of DCFL on two benchmarks at a 2\% participation rate. In particular, recent methods still suffer a significant performance drop compared to FedAvg due to lower participation rates. Despite these challenges, DCFL demonstrates promising performance gains by independently adjusting the attraction and repulsion forces.

%% file: main.bbl
\begin{thebibliography}{33}
\providecommand{\natexlab}[1]{#1}

\bibitem[{Awasthi, Dikkala, and Kamath(2022)}]{awasthi2022more}
Awasthi, P.; Dikkala, N.; and Kamath, P. 2022.
\newblock Do more negative samples necessarily hurt in contrastive learning?
\newblock In \emph{International conference on machine learning}, 1101--1116. PMLR.

\bibitem[{Chen et~al.(2020)Chen, Kornblith, Norouzi, and Hinton}]{chen2020simple}
Chen, T.; Kornblith, S.; Norouzi, M.; and Hinton, G. 2020.
\newblock A simple framework for contrastive learning of visual representations.
\newblock In \emph{International conference on machine learning}, 1597--1607. PmLR.

\bibitem[{Collins et~al.(2021)Collins, Hassani, Mokhtari, and Shakkottai}]{collins2021exploiting}
Collins, L.; Hassani, H.; Mokhtari, A.; and Shakkottai, S. 2021.
\newblock Exploiting shared representations for personalized federated learning.
\newblock In \emph{International conference on machine learning}, 2089--2099. PMLR.

\bibitem[{Dong and Voiculescu(2021)}]{dong2021federated}
Dong, N.; and Voiculescu, I. 2021.
\newblock Federated contrastive learning for decentralized unlabeled medical images.
\newblock In \emph{International Conference on Medical Image Computing and Computer-Assisted Intervention}, 378--387. Springer.

\bibitem[{Fu et~al.(2025)Fu, Chen, He, Wang, Zhang, Chen, and Li}]{fu2025virtual}
Fu, X.; Chen, Z.; He, Y.; Wang, S.; Zhang, B.; Chen, C.; and Li, J. 2025.
\newblock Virtual Nodes Can Help: Tackling Distribution Shifts in Federated Graph Learning.
\newblock In \emph{Proceedings of the AAAI Conference on Artificial Intelligence}, volume~39, 16657--16665.

\bibitem[{He et~al.(2020)He, Fan, Wu, Xie, and Girshick}]{he2020momentum}
He, K.; Fan, H.; Wu, Y.; Xie, S.; and Girshick, R. 2020.
\newblock Momentum contrast for unsupervised visual representation learning.
\newblock In \emph{Proceedings of the IEEE/CVF conference on computer vision and pattern recognition}, 9729--9738.

\bibitem[{He et~al.(2016)He, Zhang, Ren, and Sun}]{he2016deep}
He, K.; Zhang, X.; Ren, S.; and Sun, J. 2016.
\newblock Deep residual learning for image recognition.
\newblock In \emph{Proceedings of the IEEE conference on computer vision and pattern recognition}, 770--778.

\bibitem[{Hsu, Qi, and Brown(2019)}]{hsu2019measuring}
Hsu, T.-M.~H.; Qi, H.; and Brown, M. 2019.
\newblock Measuring the effects of non-identical data distribution for federated visual classification.
\newblock \emph{arXiv preprint arXiv:1909.06335}.

\bibitem[{Huang and Liu(2025)}]{huang2025pa3fed}
Huang, C.; and Liu, B. 2025.
\newblock Pa3fed: Period-aware adaptive aggregation for improved federated learning.
\newblock In \emph{Proceedings of the AAAI Conference on Artificial Intelligence}, volume~39, 17395--17403.

\bibitem[{Karimireddy et~al.(2020)Karimireddy, Kale, Mohri, Reddi, Stich, and Suresh}]{karimireddy2020scaffold}
Karimireddy, S.~P.; Kale, S.; Mohri, M.; Reddi, S.; Stich, S.; and Suresh, A.~T. 2020.
\newblock Scaffold: Stochastic controlled averaging for federated learning.
\newblock In \emph{International conference on machine learning}, 5132--5143. PMLR.

\bibitem[{Khosla et~al.(2020)Khosla, Teterwak, Wang, Sarna, Tian, Isola, Maschinot, Liu, and Krishnan}]{khosla2020supervised}
Khosla, P.; Teterwak, P.; Wang, C.; Sarna, A.; Tian, Y.; Isola, P.; Maschinot, A.; Liu, C.; and Krishnan, D. 2020.
\newblock Supervised contrastive learning.
\newblock \emph{Advances in neural information processing systems}, 33: 18661--18673.

\bibitem[{Kim, Kim, and Han(2024)}]{kim2024communication}
Kim, G.; Kim, J.; and Han, B. 2024.
\newblock Communication-efficient federated learning with accelerated client gradient.
\newblock In \emph{Proceedings of the IEEE/CVF Conference on Computer Vision and Pattern Recognition}, 12385--12394.

\bibitem[{Krizhevsky, Hinton et~al.(2009)}]{krizhevsky2009learning}
Krizhevsky, A.; Hinton, G.; et~al. 2009.
\newblock Learning multiple layers of features from tiny images.

\bibitem[{Le and Yang(2015)}]{le2015tiny}
Le, Y.; and Yang, X. 2015.
\newblock Tiny imagenet visual recognition challenge.
\newblock \emph{CS 231N}, 7(7): 3.

\bibitem[{Lee et~al.(2024)Lee, Jeong, Kim, Oh, and Yun}]{lee2024fedsol}
Lee, G.; Jeong, M.; Kim, S.; Oh, J.; and Yun, S.-Y. 2024.
\newblock Fedsol: Stabilized orthogonal learning with proximal restrictions in federated learning.
\newblock In \emph{Proceedings of the IEEE/CVF Conference on Computer Vision and Pattern Recognition}, 12512--12522.

\bibitem[{Li, He, and Song(2021)}]{li2021model}
Li, Q.; He, B.; and Song, D. 2021.
\newblock Model-contrastive federated learning.
\newblock In \emph{Proceedings of the IEEE/CVF conference on computer vision and pattern recognition}, 10713--10722.

\bibitem[{Li et~al.(2020)Li, Sahu, Zaheer, Sanjabi, Talwalkar, and Smith}]{li2020federated}
Li, T.; Sahu, A.~K.; Zaheer, M.; Sanjabi, M.; Talwalkar, A.; and Smith, V. 2020.
\newblock Federated optimization in heterogeneous networks.
\newblock \emph{Proceedings of Machine learning and systems}, 2: 429--450.

\bibitem[{Li et~al.(2019)Li, Huang, Yang, Wang, and Zhang}]{li2019convergence}
Li, X.; Huang, K.; Yang, W.; Wang, S.; and Zhang, Z. 2019.
\newblock On the convergence of fedavg on non-iid data.
\newblock \emph{arXiv preprint arXiv:1907.02189}.

\bibitem[{Liu et~al.(2025)Liu, Han, Li, and Liu}]{liu2025semidfl}
Liu, X.; Han, P.; Li, X.; and Liu, B. 2025.
\newblock SemiDFL: A Semi-Supervised Paradigm for Decentralized Federated Learning.
\newblock In \emph{Proceedings of the AAAI Conference on Artificial Intelligence}, volume~39, 18987--18995.

\bibitem[{McMahan et~al.(2017)McMahan, Moore, Ramage, Hampson, and y~Arcas}]{mcmahan2017communication}
McMahan, B.; Moore, E.; Ramage, D.; Hampson, S.; and y~Arcas, B.~A. 2017.
\newblock Communication-efficient learning of deep networks from decentralized data.
\newblock In \emph{Artificial intelligence and statistics}, 1273--1282. PMLR.

\bibitem[{Mendieta et~al.(2022)Mendieta, Yang, Wang, Lee, Ding, and Chen}]{mendieta2022local}
Mendieta, M.; Yang, T.; Wang, P.; Lee, M.; Ding, Z.; and Chen, C. 2022.
\newblock Local learning matters: Rethinking data heterogeneity in federated learning.
\newblock In \emph{Proceedings of the IEEE/CVF Conference on Computer Vision and Pattern Recognition}, 8397--8406.

\bibitem[{Mu et~al.(2023)Mu, Shen, Cheng, Geng, Fu, Zhang, and Zhang}]{mu2023fedproc}
Mu, X.; Shen, Y.; Cheng, K.; Geng, X.; Fu, J.; Zhang, T.; and Zhang, Z. 2023.
\newblock Fedproc: Prototypical contrastive federated learning on non-iid data.
\newblock \emph{Future Generation Computer Systems}, 143: 93--104.

\bibitem[{Reddi et~al.(2020)Reddi, Charles, Zaheer, Garrett, Rush, Kone{\v{c}}n{\`y}, Kumar, and McMahan}]{reddi2020adaptive}
Reddi, S.; Charles, Z.; Zaheer, M.; Garrett, Z.; Rush, K.; Kone{\v{c}}n{\`y}, J.; Kumar, S.; and McMahan, H.~B. 2020.
\newblock Adaptive federated optimization.
\newblock \emph{arXiv preprint arXiv:2003.00295}.

\bibitem[{Seo et~al.(2024)Seo, Kim, Kim, and Han}]{seo2024relaxed}
Seo, S.; Kim, J.; Kim, G.; and Han, B. 2024.
\newblock Relaxed contrastive learning for federated learning.
\newblock In \emph{Proceedings of the IEEE/CVF Conference on Computer Vision and Pattern Recognition}, 12279--12288.

\bibitem[{Shi et~al.(2022)Shi, Liang, Zhang, Tan, and Bai}]{shi2022towards}
Shi, Y.; Liang, J.; Zhang, W.; Tan, V.~Y.; and Bai, S. 2022.
\newblock Towards understanding and mitigating dimensional collapse in heterogeneous federated learning.
\newblock \emph{arXiv preprint arXiv:2210.00226}.

\bibitem[{Tan et~al.(2022)Tan, Long, Ma, Liu, Zhou, and Jiang}]{tan2022federated}
Tan, Y.; Long, G.; Ma, J.; Liu, L.; Zhou, T.; and Jiang, J. 2022.
\newblock Federated learning from pre-trained models: A contrastive learning approach.
\newblock \emph{Advances in neural information processing systems}, 35: 19332--19344.

\bibitem[{Tian, Krishnan, and Isola(2020)}]{tian2020contrastive}
Tian, Y.; Krishnan, D.; and Isola, P. 2020.
\newblock Contrastive multiview coding.
\newblock In \emph{European conference on computer vision}, 776--794. Springer.

\bibitem[{Wang et~al.(2020{\natexlab{a}})Wang, Yurochkin, Sun, Papailiopoulos, and Khazaeni}]{wang2020federated}
Wang, H.; Yurochkin, M.; Sun, Y.; Papailiopoulos, D.; and Khazaeni, Y. 2020{\natexlab{a}}.
\newblock Federated learning with matched averaging.
\newblock \emph{arXiv preprint arXiv:2002.06440}.

\bibitem[{Wang et~al.(2020{\natexlab{b}})Wang, Liu, Liang, Joshi, and Poor}]{wang2020tackling}
Wang, J.; Liu, Q.; Liang, H.; Joshi, G.; and Poor, H.~V. 2020{\natexlab{b}}.
\newblock Tackling the objective inconsistency problem in heterogeneous federated optimization.
\newblock \emph{Advances in neural information processing systems}, 33: 7611--7623.

\bibitem[{Wang and Isola(2020)}]{wang2020understanding}
Wang, T.; and Isola, P. 2020.
\newblock Understanding contrastive representation learning through alignment and uniformity on the hypersphere.
\newblock In \emph{International conference on machine learning}, 9929--9939. PMLR.

\bibitem[{Wu et~al.(2018)Wu, Xiong, Yu, and Lin}]{wu2018unsupervised}
Wu, Z.; Xiong, Y.; Yu, S.~X.; and Lin, D. 2018.
\newblock Unsupervised feature learning via non-parametric instance discrimination.
\newblock In \emph{Proceedings of the IEEE conference on computer vision and pattern recognition}, 3733--3742.

\bibitem[{Xia et~al.(2025)Xia, Hu, Yan, Liu, Li, Xie, and Chen}]{xia2025multisfl}
Xia, Z.; Hu, M.; Yan, D.; Liu, R.; Li, A.; Xie, X.; and Chen, M. 2025.
\newblock MultiSFL: Towards Accurate Split Federated Learning via Multi-Model Aggregation and Knowledge Replay.
\newblock In \emph{Proceedings of the AAAI Conference on Artificial Intelligence}, volume~39, 914--922.

\bibitem[{Xu et~al.(2025)Xu, Li, Wu, and Ren}]{xu2025federated}
Xu, H.; Li, J.; Wu, W.; and Ren, H. 2025.
\newblock Federated learning with sample-level client drift mitigation.
\newblock In \emph{Proceedings of the AAAI Conference on Artificial Intelligence}, volume~39, 21752--21760.

\end{thebibliography}
